# Bayesian Nonexhaustive Learning for Online Discovery and Modeling of Emerging Classes

**Murat Dundar**                                                                                  DUNDAR@CS.IUPUI.EDU
Department of Computer and Information Science, IUPUI, Indianapolis, IN 46202 USA

**Ferit Akova**                                                                                   FERAKOVA@CS.IUPUI.EDU
Department of Computer and Information Science, IUPUI, Indianapolis, IN 46202 USA

**Yuan Qi**                                                                                       ALANQI@CS.PURDUE.EDU
Computer Science Department, Purdue University, W. Lafayette, IN 47906 USA

**Bartek Rajwa**                                                                                  BRAJWA@PURDUE.EDU
Biindley Bioscience Center, Purdue University, W. Lafayette, IN 47906 USA

## Abstract

We present a framework for online inference in the presence of a nonexhaustively defined set of classes that incorporates supervised classification with class discovery and modeling. A Dirichlet process prior (DPP) model defined over class distributions ensures that both known and unknown class distributions originate according to a common base distribution. In an attempt to automatically discover potentially interesting class formations, the prior model is coupled with a suitably chosen data model, and sequential Monte Carlo sampling is used to perform online inference. Our research is driven by a biodetection application, where a new class of pathogen may suddenly appear, and the rapid increase in the number of samples originating from this class indicates the onset of an outbreak.

## 1. Introduction

A training set is considered *exhaustive* if it contains samples from all classes of informational value. When some classes are missing and hence not represented, the resulting training set is considered *nonexhaustive*. It is impractical, often impossible, to define a training set with a complete set of classes and then collect samples for each class, mainly because some of the classes may not be in existence at the time of training, they may exist but are not known, or their existence may be known but samples are simply not obtainable. A traditional supervised classifier trained using a nonexhaustive training set misclassifies a sample of a missing class with a probability one, making the associated learning problem ill-defined.

### 1.1. Motivation

The current research is driven mainly by a biosensing problem involving prediction of the presence of specific as well as unmatched/emerging pathogenic microorganisms in various biological samples. A global surge in the number of outbreaks together with elevated concerns about biosecurity has led to an enormous interest among scientific communities and government agencies in developing reagantless techniques for rapid identification of pathogens. Traditional recognition methods based on antibodies or genetic matching remain labor intensive and time consuming, and involve multiple steps. Recent studies based on quantitative phenotypic evaluation has shown great promise for distinguishing bacterial cultures at the genus, species, and strain level.

The core advantage of label-free methods is their ability to quantify phenotypes for which there are no available antibodies or genetic markers. This information can be used within a traditional supervised-learning framework in which knowledge discovered from independently tested and prelabeled samples is used for training. However, the quality of training libraries is potentially limited because the sheer number of bacterial classes would not allow for practical and manageable training in a traditional supervised setting; for instance *Salmonella* alone has over 2400 known serovars. Additionally, microorganisms are characterized by a high mutation rate, which indicates new classes of bacteria can emerge anytime. Nonexhaustive learning when implemented in this domain in an online fashion aims at rapid identification of new, emerging classes of microorganisms, which are not represented in the initial training library. Ability to detect the sudden presence of a new class (or classes) would be an important element of an automated outbreak-identification strategy.

### 1.2. Proposed approach in a nutshell

Using a nonexhaustive training dataset, a Dirichlet process prior (DPP) is coupled with a Normal data model to deal with known as well as unknown classes. The parameters of the base distribution, which is chosen as a bivariate Normal × Inverted Wishart distribution, are estimated using samples initially available for known classes. A sequential importance resampling (SIR) technique is proposed to perform online inference to efficiently evaluate the probability of a new sample belonging to an emerging class or one of the existing ones without the need for explicit knowledge of the class labels of previously observed samples. In this framework new classes characterized by a rapid increase in sample size is of significance for early identification of potentially interesting class formations.

### 1.3. Related Work

Early work most similar to nonexhaustive learning includes the two studies reported in (Akova et al., 2010; Miller & Browning, 2003). In (Akova et al., 2010) a Bayesian approach based on the maximum likelihood detection of novelties using a dynamically updated class set was proposed. In (Miller & Browning, 2003) known and unknown classes were modeled by a mixture of experts model with learning performed by expectation-maximization. The former depends on the class conditional likelihoods for creating new classes and the latter uses minimum description length coupled with some heuristics to determine the optimal number of mixture components. Neither of the approaches consider a prior model for class distributions which results in the decision to create a new class to be mainly data driven and ad-hoc in both approaches. Also the lack of efficient online/incremental learning capabilities makes both approaches impractical for processing large sequential data.

Other work related to nonexhaustive learning can be reviewed within the scope of offline anomaly/novelty detection, online/incremental learning, and online clustering with novelty detection. Most of the early work on offline anomaly/novelty detection is developed around one-class classification problems and uses either support estimation, one-class classifiers, or density-based models to identify novelties. These techniques provide an offline framework for detecting novelties but do not differentiate among them; thus, lack the capability to discover and model individual classes online.

Online/incremental learning develops efficient algorithms for sequential classification problems such that the classifier model can be updated using only the current sample without retraining with past samples. Many of these studies assume that the initial training set is exhaustive. The one (Kivinen et al., 2001) that does consider nonexhaustiveness studies novelty detection along with online learning, but its scope is limited to one class problems only.

Another line of work related to nonexhaustive learning has been developed in the area of online clustering with and without novelty detection. We will use DPP in this study for online class modeling the same way these techniques use it for online cluster modeling. However, unlike online cluster modeling, which depends on a vague prior, a more informative prior can be obtained in a nonexhaustive setting using samples of represented classes. To avoid the vague prior issue the work in (Zhang et al., 2005) proposes using a historical dataset to estimate the parameters of the DPP. Although the work is framed as a clustering problem with a multinomial data model, it is similar to the proposed study in using labeled data for estimating the prior model. However, the solution offered in this approach for the sequential update of the cluster models is suboptimal in that the posterior probability for each incoming sample was evaluated once and same values were used for all subsequent samples. As new samples emerge these previously evaluated probability values can increase/decrease resulting in suboptimal class assignments for samples observed earlier. We believe that the proposed sequential inference technique involving particle filters is an important step toward addressing this problem in a nonexhaustive setting.

## 2. Problem formulation

In this section we present a general framework for learning with a nonexhaustively-defined training dataset, which allows for online discovery as well as modeling of new classes. To differentiate classes discovered online from those initially available in the training library we introduce the notion of *labeled* vs. *unlabeled* classes, where the terms labeled and unlabeled refer to verified and unverified classes, respectively. When referring to classes discovered online the terms *unlabeled class* and *cluster* are used interchangeably throughout this text. In the proposed framework online class modeling is tackled by a DPP model (Ferguson, 1973).

## 2.1. Dirichlet process prior

Let $x_i$, $i = \{1, \ldots, n\}$ be the feature vector characterizing a sample in the d-dimensional vector space $\Re$ and $y_i$ be its corresponding class indicator variable. If $x_i$ is distributed according to an unknown distribution $p(.|\theta_i)$, then defining a DPP over class distributions is equivalent to modeling the prior distribution of $\theta$ by a Dirichlet process. More formally,

$$\begin{aligned} x_i|\theta_i &\sim p(\cdot|\theta_i) \\ \theta_i &\sim G(\cdot) \\ G &\sim DP(\cdot|G_0, \alpha) \end{aligned} \quad (1)$$

where $G$ is a random probability measure, which is distributed according to a Dirichlet process (DP) defined by a base distribution, $G_0$, and the precision parameter, $\alpha$. Given that $G$ is distributed according to a DP, the stick-breaking construction due to (Ishwaran & James, 2001) suggests $G = \sum_{i=1}^{\infty} \beta_i \delta_{\phi_i}$ where $\beta_i = \beta'_i \prod_{l=1}^{i-1}(1 - \beta'_l)$, $\beta'_i \sim Beta(1, \alpha)$, and $\phi_i \sim G_0$. The points $\phi_i$ are called the *atoms* of $G$. Note that unlike continous distributions the probability of sampling the same $\phi_i$ twice is not zero and proportional to $\beta_i$. Thus, $G$ is considered a discrete distribution.

## 2.2. DPP in a nonexhaustive framework

The suitability of the DPP model for nonexhaustive learning can be better conceived with the help of the conditional prior of $\theta$. Let's assume that at a certain time point the training set contains a sequence of $n$ samples. The conditional prior of $\theta_{n+1}$ conditioned on all past $\theta_i$, $i = \{1, \ldots, n\}$ can be obtained by integrating out $G$ in (1) which becomes,

$$\theta_{n+1}|\theta_1, \ldots, \theta_n \sim \frac{\alpha}{\alpha + n} G_0(\cdot) + \frac{1}{\alpha + n} \sum_{i=1}^{n} \delta_{\theta_i} \quad (2)$$

This conditional prior can be interpreted as a mixture of two distributions. Any sample that originates from this prior comes from the base distribution $G_0(\cdot)$ with a probability of $\frac{\alpha}{\alpha+n}$ or uniformly generated from $\{\theta_1, \ldots, \theta_n\}$ with a probability of $\frac{n}{\alpha+n}$. With a positive probability a sequence of $n$ samples generated this way will not be all distinct. If we assume that there are $k \leq n$ distinct values of $\theta$ in a sequence of size $n$, then (2) can be rewritten,

$$\theta_{n+1}|\theta_1, \ldots, \theta_n \sim \frac{\alpha}{\alpha + n} G_0(\cdot) + \frac{1}{\alpha + n} \sum_{j=1}^{k} n_j \delta_{\theta_j^*} \quad (3)$$

where $\theta_j^*$, $j = \{1, \ldots, k\}$ are the distinct values of $\theta_i$ and $n_j$ are the number of occurrences of each $\theta_j^*$ in the sequence. Each $\theta_j^*$ defines a unique class with an indicator variable $y_j^*$, whose samples are distributed according to the probability distribution $p(\cdot|\theta_j^*)$. Based on (3), after a sequence of $n$ samples are generated, $y_{n+1} = y_j^*$ with probability equal to $\frac{n_j}{\alpha+n}$, and $y_{n+1} = y_{k+1}^*$, with probability equal to $\frac{\alpha}{\alpha+n}$, where $y_{k+1}^*$ is the new class whose parameter is defined by $\theta_{k+1}^*$ and sampled from $G_0(\cdot)$.

This prior model can also be illustrated as a *Chinese Restaurant process* (CRP) (Aldous, 1985). The CRP uses a metaphor of a Chinese restaurant with infinitely many tables where the $(n + 1)^{th}$ customer sits at a previously occupied table $j$ with a probability of $\frac{n_j}{\alpha+n}$ and at a new table $k + 1$ with a probability of $\frac{\alpha}{\alpha+n}$. Here $n_j$ is the number of customers sitting at table $j$ and $n$ is the total number of customers.

Our discussion so far has been limited to the prior model. Next, we will incorporate the data model and use the conditional posterior to determine whether a new sample $x_{n+1}$ should be assigned to one of the existing classes or to a new class sampled from $G_0$. More specifically, we are interested in the distribution $p(\theta_{n+1}|x_{n+1}, \theta_1, \ldots, \theta_n)$, which is proportional to

$$\begin{aligned} &p(\theta_{n+1}|x_{n+1}, \theta_1, \ldots, \theta_n) \\ &\propto \frac{\alpha}{\alpha+n} p(x_{n+1}) p(\theta_{n+1}|x_{n+1}) \\ &+ \frac{1}{\alpha+n} \sum_{j=1}^{k} n_j p(x_{n+1}|\theta_j^*) \delta_{\theta_j^*} \end{aligned} \quad (4)$$

which indicates $x_{n+1}$ either comes from a new class, $y_{k+1}^*$, which inherits $\theta_{k+1}^*$ sampled from $p(\theta_{n+1}|x_{n+1})$, with a probability proportional to $\frac{\alpha}{\alpha+n} p(x_{n+1})$ or belongs to $y_j^*$ with a probability proportional to $\frac{n_j}{\alpha+n} p(x_{n+1}|\theta_j^*)$.

Since $\theta_j^*$ are not known and has to be estimated using samples in the represented classes, $p(x_{n+1}|\theta_j^*)$ can be replaced with the class conditional predictive distribution $p(x_{n+1}|D_j)$ where $D_j = \{x_i\}_{i \in C^j}$ denotes the subset of samples belonging to class $y_j^*$ defined by the index set $C^j$. Thus, provided that class membership information for all samples processed before $x_{n+1}$ are known, the decision function to assign $x_{n+1}$ to a new class or one of the existing ones can be expressed as,

$$h(x_{n+1}) = \begin{cases} y_{n+1} = y_j^* & \text{if} \\ \frac{n_{j^*}}{\alpha+n} p(x_{n+1}|D_{j^*}) \geqslant \frac{\alpha}{\alpha+n} p(x_{n+1}) \\ y_{n+1} = y_{k+1}^* & \text{if} \\ \frac{n_{j^*}}{\alpha+n} p(x_{n+1}|D_{j^*}) < \frac{\alpha}{\alpha+n} p(x_{n+1}) \end{cases} \quad (5)$$

where $j^* = \mathrm{argmax}_j \left\{ \frac{n_j}{\alpha+n} p(x_{n+1}|D_j) \right\}_{j=1}^{k}$. However, in the nonexhaustive learning framework class mem-bership information is only available for samples initially present in the training dataset. For all samples processed online before $x_{n+1}^{th}$ sample the true class membership information is unknown.

## 3. Inference with a nonexhaustive set of classes

Before we move on to discussing how inference can be performed in this framework, we introduce new notation to distinguish between the two types of samples available during online execution: samples initially available in the training dataset with known class membership information and samples observed online with no verified class membership information. Let $X = \{x_1, \ldots, x_\ell\}$ be the set of all training samples initially available, $Y = \{y_1, \ldots, y_\ell\}$ be the corresponding set of known class indicator variables with $y_i \in \{1, \ldots, k\}$, $k$ being the number of known classes, $\tilde{X}^n = \{\tilde{x}_1, \ldots, \tilde{x}_n\}$ be the set of $n$ samples sequentially observed online, and $\tilde{Y}^n = \{\tilde{y}_1, \ldots, \tilde{y}_n\}$ be the corresponding set of unknown class indicator variables with $\tilde{y}_i \in \{1, \ldots, \tilde{k} + k\}$, $\tilde{k}$ being the number of unrepresented classes associated with these $n$ samples.

### 3.1. Inference by Gibbs Sampling

We are interested in predicting $\tilde{Y}_{n+1}$, i.e., the class labels for all $\tilde{X}_{n+1}$ at the time $\tilde{x}_{n+1}$ is observed, which can be done by finding the mean of the posterior distribution $p(\tilde{Y}^{n+1}|\tilde{X}^{n+1}, X, Y)$. Although this integral cannot be easily evaluated, the closed form solution for the conditional distributions of the latent variables $\tilde{y}_i$ can easily be obtained. Thus, Gibbs sampling with the sampler state consisting of variables $\tilde{y}_i$, $i = \{1, \ldots, n+1\}$, can be used to approximate $p(\tilde{Y}^{n+1}|\tilde{X}^{n+1}, X, Y)$. One sweep of the Gibbs sampler will involve sampling from the following conditional distribution $\forall i$.

$$\begin{aligned} &p(\tilde{y}_i|\tilde{Y}^{(n+1)/i}, \tilde{X}^{n+1}, X, Y) \\ &\propto \frac{\alpha}{\alpha+n+\ell} p(\tilde{x}_i) \delta_{\tilde{k}+k+1} \\ &+ \frac{1}{\alpha+n+\ell} \sum_{j=1}^{k+\tilde{k}} n_j p(\tilde{x}_i|D_j) \delta_j \end{aligned} \quad (6)$$

where $\tilde{Y}^{(n+1)/i}$ denotes $\tilde{Y}^{(n+1)}$ without $\tilde{y}_i$.

### 3.2. Inference by Sequential Importance Resampling (SIR)

With the Gibbs sampler approach every time a new sample is observed the sampler has to run from start to predict whether the current sample belongs to one of the existing classes (labeled/unlabeled) or to a new class. This sampling scheme eventually becomes intractable as the number of unlabeled samples gradually increases. We believe that this problem can be addressed to a greater extent by developing a sequential sampling approach based on Sequential Importance Resampling (SIR) (Doucet et al., 2000). In this approach, at any given time, the sampler only depends on a set of particles and their corresponding weights, which are efficiently updated in a sequential manner each time a new sample is observed without the need for the past samples.

More specifically, we are interested in evaluating the expectation $E_{p(\tilde{Y}^{n+1}|\tilde{Y}^n, \tilde{X}^{n+1}, Y, X)}\left[\tilde{Y}^{n+1}\right]$. Using an importance function $q(\tilde{Y}^{n+1}|\tilde{Y}^n, \tilde{X}^{n+1}, Y, X)$ the relevant integral can be approximated as follows.

$$\begin{aligned} &E_{p(\tilde{Y}^{n+1}|\tilde{Y}^n, \tilde{X}^{n+1}, Y, X)}\left[\tilde{Y}^{n+1}\right] \\ &= \int \tilde{Y}^{n+1} p(\tilde{Y}^{n+1}|\tilde{Y}^n, \tilde{X}^{n+1}, Y, X) \partial \tilde{Y}^{n+1} \\ &= \int \tilde{Y}^{n+1} w_{n+1}(\tilde{Y}^{n+1}) q(\tilde{Y}^{n+1}|\tilde{Y}^n, \tilde{X}^{n+1}, Y, X) \partial \tilde{Y}^{n+1} \\ &\approx \sum_{m=1}^{M} \tilde{Y}^{n+1} w_m^{n+1}(\tilde{Y}^{n+1}) \delta_{\tilde{Y}_m^{n+1}} \end{aligned} \quad (7)$$

where $M$ is the number of particles and $w_m^{n+1}(\tilde{Y}^{n+1}) = \frac{p(\tilde{Y}^{n+1}|\tilde{Y}^n, \tilde{X}^{n+1}, Y, X)}{q(\tilde{Y}^{n+1}|\tilde{Y}^n, \tilde{X}^{n+1}, Y, X)}$ is the corresponding weight of the $m^{th}$ particle at the time $(n+1)^{th}$ sample is observed. Particles are sampled from the importance function $q(\tilde{Y}^{n+1}|\tilde{Y}^n, \tilde{X}^{n+1}, Y, X)$. The weights can be evaluated sequentially up to an unknown constant as outlined next.

Using the chain rule and after some manipulations a sequential update formula for the particle weights $w_{n+1}^m(\tilde{Y}^{n+1})$ can be derived as follows

$$\begin{aligned} &w_m^{n+1}(\tilde{Y}^{n+1}) \\ &= \frac{p(\tilde{Y}^{n+1}|\tilde{Y}^n, \tilde{X}^{n+1}, Y, X)}{q(\tilde{Y}^{n+1}|\tilde{Y}^n, \tilde{X}^{n+1}, Y, X)} \\ &= w_m^n(\tilde{Y}^n) \frac{p(\tilde{x}_{n+1}|\tilde{Y}^{n+1}, \tilde{X}^n, Y, X) p(\tilde{y}_{n+1}|\tilde{Y}^n, Y)}{p(\tilde{x}_{n+1}|\tilde{Y}^n, \tilde{X}^n, Y, X) q(\tilde{y}_{n+1}|\tilde{Y}^n, \tilde{X}^{n+1}, Y, X)} \end{aligned} \quad (8)$$

Although, it is not optimal in terms of minimizing the variance, the common choice for $q(\tilde{y}_{n+1}|\tilde{Y}^n, \tilde{X}^{n+1}, Y, X) = p(\tilde{y}_{n+1}|\tilde{Y}^n, Y)$ further simplifies the update formula by canceling out both terms in (8). After considering the fact that $p(\tilde{x}_{n+1}|\tilde{Y}^n, \tilde{X}^n, Y, X)$ is constant with respect to $\tilde{Y}^{n+1}$, the sequential update formula for the particle weights become

$$w_m^{n+1}(\tilde{Y}^{n+1}) = C w_m^n(\tilde{Y}^n) p(\tilde{x}_{n+1}|\tilde{Y}^{n+1}, \tilde{X}^n, Y, X) \quad (9)$$

Since $p(\tilde{x}_{n+1}|\tilde{Y}^{n+1}, \tilde{X}^n, Y, X)$ can be evaluated for $\tilde{x}_{n+1}$ for any given particle, the weights at stage $n+1$ can be obtained up to an unknown constant $C$. Using normalized weights eliminates $C$ and thus the discrete probability distribution in (7) can be fully evaluated and efficiently updated.

Every time a new sample is observed, first, a designated number of, i.e., $R$, new particles are resampled for each of the $M$ particle using the importance function, then, weights are updated for the $M*R$ particles, finally, downsampling, stratified on the particle weights, is performed to select $M$ particles out of $M*R$ ones. Resampling is critical to avoid the weight degeneracy problem mainly associated with Dirichlet process mixture models. To address the weight degeneracy problem a resampling strategy that ensures a well-distributed particle set was introduced in (Fearnhead & Clifford, 2003; Wood & Black, 2008). Using this strategy, instead of resampling $R$ new particles for each existing particle from the importance function, $k + \tilde{k} + 1$ particles are generated by considering all possible class labels an incoming sample can take for a given particle. Note that although $k$, i.e., number of labeled classes is constant across all particles, $\tilde{k}$, i.e., number of unlabeled classes, varies from one particle to other. Since all possible classes are considered in this approach, it is now essential to revise the weight update formula to include prior probability for each class.

### 3.3. Estimating the precision parameter $\alpha$

In the proposed framework $\alpha$ is the parameter that controls the prior probability of assigning a new sample to a new cluster and thus, plays a critical role in the number of clusters generated. When the training samples are collected to reflect the true proportion of each class as well as the actual number of classes as in a training set with the same number of samples collected in real-time, the marginal distribution of the number of clusters $p(\tilde{k})$ can be maximized to obtain the maximum likelihood estimate of $\alpha$. However, in many machine learning applications only the most prevalent classes are available in the training set and the training samples are almost never collected in real-time. Thus, $p(\tilde{k})$ may not model a training set collected offline very well.

One viable approach to predicting $\alpha$ when training samples are not collected in real-time is to sample it from the distribution $p(\alpha|\tilde{k}, n)$ (Escobar & West, 1994). This approach although widely used in mixture density estimation involving batch data as part of a Gibbs sampler, it is not suitable for the proposed SIR algorithm, mainly because with SIR particles themselves are a function of $\alpha$. Therefore, $\alpha$ has to be fixed in order for the weight update formula to hold and thus, the SIR algorithm to work. In this study we encode our prior belief about the odds of encountering a new class by a prior probability value $p$ that indicates the prior probability of a new sample coming from one of the labeled classes in the training set. Once a vague value for $p$ is obtained for a given domain, $\alpha$ can be estimated by empirical Bayes by sampling a large number of samples from a CRP for varying values of $\alpha$ and then picking up the one that minimizes the difference between the empirical and true values of $p$.

## 4. A Normally distributed data model

Both the Gibbs sampler and SIR requires the evaluation of the predictive $p(\tilde{x}_i|D_j)$ and the marginal $p(\tilde{x}_i)$ distributions. The predictive distribution for both labeled and unlabeled classes can be obtained by integrating out $\theta$. The marginal distribution can be obtained from $p(\tilde{x}_i|D_j)$ by setting $D_j$ an empty set. In general the exact solution for the predictive and marginal distributions does not exist and approximations are needed. However, a closed-form solution does exist for a Normally distributed data model and a properly chosen base distribution as presented next. We give $\omega_j$ a Gaussian distribution with mean $\mu_j$ and covariance $\Sigma_j$; that is, $\omega_j \sim \mathcal{N}(\mu_j, \Sigma_j)$. For the mean and covariance matrix, we use a joint conjugate prior $G_0$:

$$G_0 = p(\mu, \Sigma) = \underbrace{\mathcal{N}\left(\mu|\mu_0, \frac{\Sigma}{\kappa}\right)}_{p(\mu|\Sigma)} \times \underbrace{W^{-1}(\Sigma|\Sigma_0, m)}_{p(\Sigma)} \quad (10)$$

where $\mu_0$ is the prior mean and $\kappa$ is a scaling constant that controls the deviation of the class conditional mean vectors from the prior mean. The smaller the $\kappa$ is, the larger the between class scattering will be. The parameter $\Sigma_0$ is a positive definite matrix that encodes our prior belief about the expected $\Sigma$. The parameter $m$ is a scalar that is negatively correlated with the degrees of freedom. In other words the larger the $m$ is the less $\Sigma$ will deviate from $\Sigma_0$ and vice versa.

To evaluate the update formula in (9) for SIR we need $p(x_{n+1}|D_j)$. To obtain $p(x_{n+1}|D_j)$ we need to integrate out $\theta = \{\mu, \Sigma\}$. Since the sample mean $\bar{x}$ and the sample covariance matrix $S$ are sufficient statistics for the multivariate Normally distributed data, we can write $p(\mu, \Sigma|D_j) = p(\mu, \Sigma|\bar{x}_j, S_j)$. The formula for this posterior and its derivation is widely available in books on multivariate statistics (Anderson, 2003). Once we integrate out $p(x_{n+1}, \mu, \Sigma|\bar{x}_j, S_j)$ first with respect to $\mu$ and then with respect to $\Sigma$ we obtain the predictive distribution in the form of a multivariate Student-t distribution.

In addition to $p(x_{n+1}|D_j)$ we also need $p(x_{n+1})$ when evaluating the decision function in (5), which is also a multivariate Student-t distribution with $D_j$ an empty set.

### 4.1. Estimating the parameters of the prior model

The parameters $(\Sigma_0, m, \mu_0, \kappa)$ of the prior model can be estimated offline using samples from the well-defined classes. The maximum-likelihood estimates for $\Sigma_0$ and $m$ do not exist. The study in (Greene & Rayens, 1989) suggests estimating $\Sigma_0$ by the unbiased and consistent estimate $S_p$, i.e., the pooled covariance, and maximizing the marginal likelihood of $(n_j - 1)S_j$ for $m > d+1$ numerically to estimate $m$. Here, $S_p$ is the pooled covariance matrix defined by

$$S_p = \frac{(m-d-1)\sum_{j=1}^{k}(n_j-1)S_j}{n-k} \quad (11)$$

where $n$ is the total number of samples in the training set, i.e., $n = \sum_{j=1}^{k} n_j$. The marginal distribution of $(n_j - 1)S_j$ can be obtained by integrating out the joint distribution $p((n_j - 1)S_j, \Sigma_j) = p((n_j-1)S_j|\Sigma_j)p(\Sigma_j)$ with respect to $\Sigma_j$. For a Normal data model $p((n_j - 1)S_j|\Sigma_j)$ is a Wishart distribution with a scale matrix $\Sigma_j$ and degrees of freedom $n_j - 1$, i.e., $(n_j - 1)S_j|\Sigma_j \sim W(\Sigma_j, n_j - 1)$ and $p(\Sigma_j)$ is an inverted Wishart distribution as defined in (10). The parameters $\kappa$ and $\mu_0$ can be estimated by maximizing the joint likelihood of $\bar{x}$ and $S$, $p(\bar{x}, S)$, with respect to $\kappa$ and $\mu_0$, respectively.

## 5. Experiments

### 5.1. An illustrative example

We present an illustrative example demonstrating the proposed algorithm discovering and modeling classes with a 2-D simulated dataset. We generate twenty three classes where the class covariance matrix of each class is obtained from an inverted Wishart distribution with parameters $\Psi = 10I$ and $m = 20$ and mean vectors are equidistantly placed alongside the peripheries of two circles with radius 4 and 8 creating a flower-shaped dataset. Here, $I$ denotes the 2-D identity matrix. Three of the twenty three classes are randomly chosen as unrepresented. The nonexhaustive training data contains twenty classes with each class represented by 100 samples (a total of 2000 samples) whereas the exhaustive testing data contains twenty three classes with 100 samples from each (a total of 2300 samples). The objective here is to discover and model the three unrepresented classes while making sure samples of represented classes are classified as accurately as possible. Figure 1a shows true class distributions for all twenty three classes. The represented classes are shown by solid lines and unrepresented ones by dashed lines. The ellipses correspond to the distributions of the classes that are at most three standard deviations away from the mean. The testing samples are classified sequentially using the SIR algorithm discussed in Section 3.2. The precision parameter $\alpha$ and the number of particles $M$ are chosen as 1 and 500, respectively. Figures 1b, 1c, and 1d demonstrate the online discovery and modeling of new classes when 100, 300, and all 2300 test samples are classified, respectively. The discovered classes are marked by solid blue lines. All three classes are discovered and their underlying distributions are successfully recovered by generating one cluster for each class.

### 5.2. Bacteria detection

A total of 2054 samples from 28 classes each representing a different bacteria serovar were considered in this study. These are the type of serovars most commonly found in food samples. Each serovar is represented by between 40 to 100 samples where samples are the *forward-scatter patterns* characterizing the phenotype of a bacterial colony obtained by illuminating the colony surface by a laser light. Each scatter pattern is a gray level image characterized by a set of 50 features. More information about this dataset is available in (Akova et al., 2010). Samples are randomly splitted into two as train and test, with 70% of the samples going into the training set and the remaining 30% in the test. Stratified sampling is used to make sure each class is proportionately represented in both the training and the test sets. Four of the classes are considered unknown and all of their samples are moved from the training set to the test set. The nonexhaustive training set contains 24 classes whereas the exhaustive testing set contains 28 classes. Since the training samples are collected offline the number of samples initially available for labeled classes may not necessarily reflect true class proportions. To avoid introducing bias in favor of classes with larger numbers of samples we assumed that each labeled class is a priori likely by setting $n_j = 1$.

The performance of the proposed SIR algorithm (NEL-SIR) discussed in Section 3.2 is evaluated on three fronts: classification accuracy for represented classes, classification accuracy for unrepresented classes, and the number of clusters discovered for each of the unrepresented classes. To compute the latter two values each unlabeled cluster is assigned to the unrepresented class having the majority of the samples in that class. Classification accuracy for each unrepresented class is computed by the ratio of the total number of samples recovered by the corresponding clusters to the total number of samples in that class. To see the effect of the execution order of the test samples on the overall results the experiment is repeated multiple times each time with a different ordering of test samples. Using the approach discussed in Section 3.3 the precision parameter $\alpha$ is predicted as 10 for a $p$ of 0.95, i.e., an incoming sample *a priori* belongs to one of the 24 labeled classes by a probability of 0.95. The number of particles $M$ is chosen to be 2000.

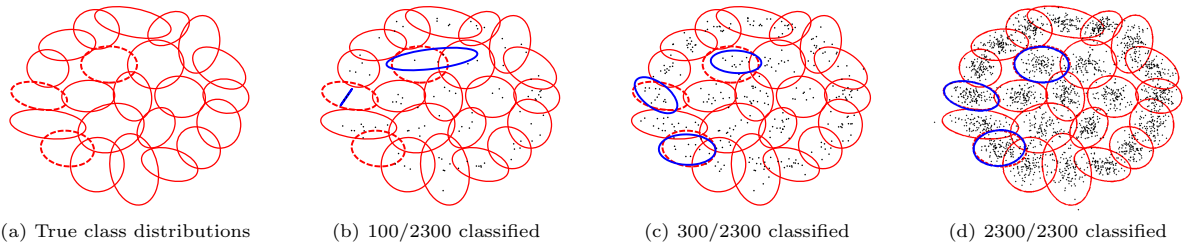

(a) True class distributions  (b) 100/2300 classified  (c) 300/2300 classified  (d) 2300/2300 classified

*Figure 1.* Illustration of the proposed algorithm with an artificial dataset. (a) Red dashed lines indicate unrepresented classes. Red solid lines indicate represented classes. (b)-(d) Blue solid lines indicate newly discovered classes. Black '·' marks indicate testing samples.

The proposed NEL-SIR is compared against the Bayes-NoDe algorithm proposed in (Akova et al., 2010). Both the proposed NEL-SIR and Bayes-NoDe use the same data model, i.e., $\omega_j \sim \mathcal{N}(\mu_j, \Sigma_j), (\mu, \Sigma) \sim \mathcal{N}\left(\mu|\mu_0, \frac{\Sigma}{\kappa}\right) \times W^{-1}\left(\Sigma|\Sigma_0, m\right)$. However, as briefly mentioned in Section 1.3, there are significant differences between the two approaches in terms of prior modeling of class distributions and performing inference in a nonexhaustive setting. In addition to these two algorithms we also considered the exhaustive case, i.e., the setting under which all 28 classes are represented in the training set to serve as a benchmark for comparing our results. The results including the classification accuracies for represented as well as unrepresented classes and the number of clusters discovered for each unrepresented class are shown in Table 1.

Classification accuracy achieved by the proposed NEL-SIR algorithm for represented classes is on par with that achieved by Bayes-NoDe. Although all four unrepresented classes are successfully discovered by both NEL-SIR and Bayes-Node, NEL-SIR tend to generate far less number of clusters in general and achieves significantly higher accuracy than BayesNoDe for each unrepresented class. The average number of clusters discovered for each unrepresented class is especially important for practical purposes because in a real-time biodetection system once new clusters are discovered as unknowns, their samples has to be analyzed offline to assess the pathogenic nature of these clusters. The less the number of clusters discovered for each unrepresented class the less time and resources offline analysis will require.

The perfect accuracies achieved for two of the unrepresented classes in the exhaustive setting indicate these classes are well separated from other classes. For the two well-separated classes NEL-SIR performs equally well with the exhaustive case. These results indicate that if the unrepresented classes are well separated the proposed approach not only discover these classes and recover them by a reasonable number of clusters but also classify their samples at an accuracy comparable to an exhaustive classifier. If the unrepresented classes are not perfectly separated these classes can still be discovered but some loss in classifier accuracy as compared to an exhaustive classifier is inevitable.

## 6. Conclusions

Online class discovery is an important problem that finds its place in many real-life applications involving evolving datasets. In this study, in an effort to discover emerging classes, a sequential inference algorithm based on particle filters was proposed for a Dirichlet process mixture model. In this approach the posterior distribution of the class indicator variables is approximated by a discrete distribution expressed by a set of particles and the corresponding weights. The particles and their weights are efficiently updated each time a new sample is observed. This way the posterior distribution is updated in a sequential manner without the need to have access to past samples enabling efficient online inference in a nonexhaustive setting. Our algorithm is validated using a 28-class bacteria with four of the classes considered unknown and promising results are obtained with respect to classification accuracy and class discovery.

The data model used in this study was limited with the Normal model. The proposed approach can be extended to problems involving more flexible class distributions by choosing a mixture model for each class data and a hierarchical DPP model over class distributions. Additionally, the model used in our studies does not explicitly model variation in class size as a function of time. Modeling time can be essential for modeling burstiness. We believe a time-dependent Dirichlet process can be useful toward achieving this end. Owing to the long tail behavior of DPP, with the current approach the probability of discovering a new class will converge to zero as $n$ goes to infinity. Although we cannot verify whether Zipf's law holds for bacteria population we believe a Pitman-Yor process can offer more control over tail behavior of the prior model.

| | | represented classes (overall accuracy) | unrepresented classes | | | |
|---|---|---|---|---|---|---|
| | | | 1 | 2 | 3 | 4 |
| Exhaustive case | accuracy (%) | 94.0 | 88.5 | 94.1 | 100.0 | 100.0 |
| Bayes-NoDe | accuracy (%) | 91.2 | 60.5 | 82.8 | 87.6 | 92.5 |
| | | (0.2) | (3.6) | (13.4) | (8.1) | (3.4) |
| | avg. # of clusters | - | 2.7 | 5.5 | 1.7 | 4.0 |
| NEL-SIR | accuracy (%) | 91.1 | 75.7 | 90.7 | 99.5 | 99.5 |
| | | (0.4) | (1.2) | (8.0) | (0.5) | (0.7) |
| | avg. # of clusters | - | 2.0 | 1.3 | 1.7 | 1.2 |

Table 1. Comparing the performance of the NEL-SIR algorithm against Bayes-NoDe and the fully exhaustive case. Numbers in parenthesis indicate standard deviations across multiple runs.

## Acknowledgments

The project was supported by Grant Number 5R21AI085531-02 from the National Institute of Allergy and Infectious Diseases (NIAID). The content is solely the responsibility of the authors and does not necessarily represent the official views of NIAID or the National Institutes of Health.